\documentclass[10pt,twocolumn,letterpaper]{article}
\usepackage[pagenumbers]{cvpr} 
\usepackage{listings}

\usepackage{subcaption}

\definecolor{cvprblue}{rgb}{0.21,0.49,0.74}
\usepackage[pagebackref,breaklinks,colorlinks,allcolors=cvprblue]{hyperref}

\def\paperID{36} 
\def\confName{CVPR}
\def\confYear{2025}

\title{iNatAg: Multi-Class Classification Models Enabled by a Large-Scale Benchmark Dataset with 4.7M Images of 2,959 Crop and Weed Species}

\author{
Naitik Jain\textsuperscript{1}\qquad
Amogh Joshi\textsuperscript{1,2,3}\qquad
Mason Earles\textsuperscript{1,3}\\[1ex]
\textsuperscript{1}University of California, Davis\\
\textsuperscript{2}Princeton University\\
\textsuperscript{3}AI Institute for Food Systems\\[1ex]
\texttt{\small ndjain@ucdavis.edu, amoghjoshi@princeton.edu, jmearles@ucdavis.edu}
}

\begin{document}
\maketitle
\begin{abstract}

Accurate identification of crop and weed species is critical for precision agriculture and sustainable farming. However, it remains a challenging task due to a variety of factors -- a high degree of visual similarity among species, environmental variability, and a continued lack of large, agriculture-specific image data. We introduce iNatAg, a large-scale image dataset which contains over 4.7 million images of 2,959 distinct crop and weed species, with precise annotations along the taxonomic hierarchy from binary crop/weed labels to specific species labels. Curated from the broader iNaturalist database, iNatAg contains data from every continent and accurately reflects the variability of natural image captures and environments. Enabled by this data, we train benchmark models built upon the Swin Transformer architecture and evaluate the impact of various modifications such as the incorporation of geospatial data and LoRA finetuning. Our best models achieve state-of-the-art performance across all taxonomic classification tasks, achieving 92.38\% on crop and weed classification. Furthermore, the scale of our dataset enables us to explore incorrect misclassifications and unlock new analytic possiblities for plant species. By combining large-scale species coverage, multi-task labels, and geographic diversity, iNatAg provides a new foundation for building robust, geolocation-aware agricultural classification systems. We release the iNatAg dataset publicly through AgML\footnote{\url{https://github.com/Project-AgML/AgML}}, enabling direct access and integration into agricultural machine learning workflows.

\end{abstract}    
\section{Introduction}

\begin{figure*}[t] 
    \centering
    \includegraphics[width=\textwidth,height=\paperheight,keepaspectratio]{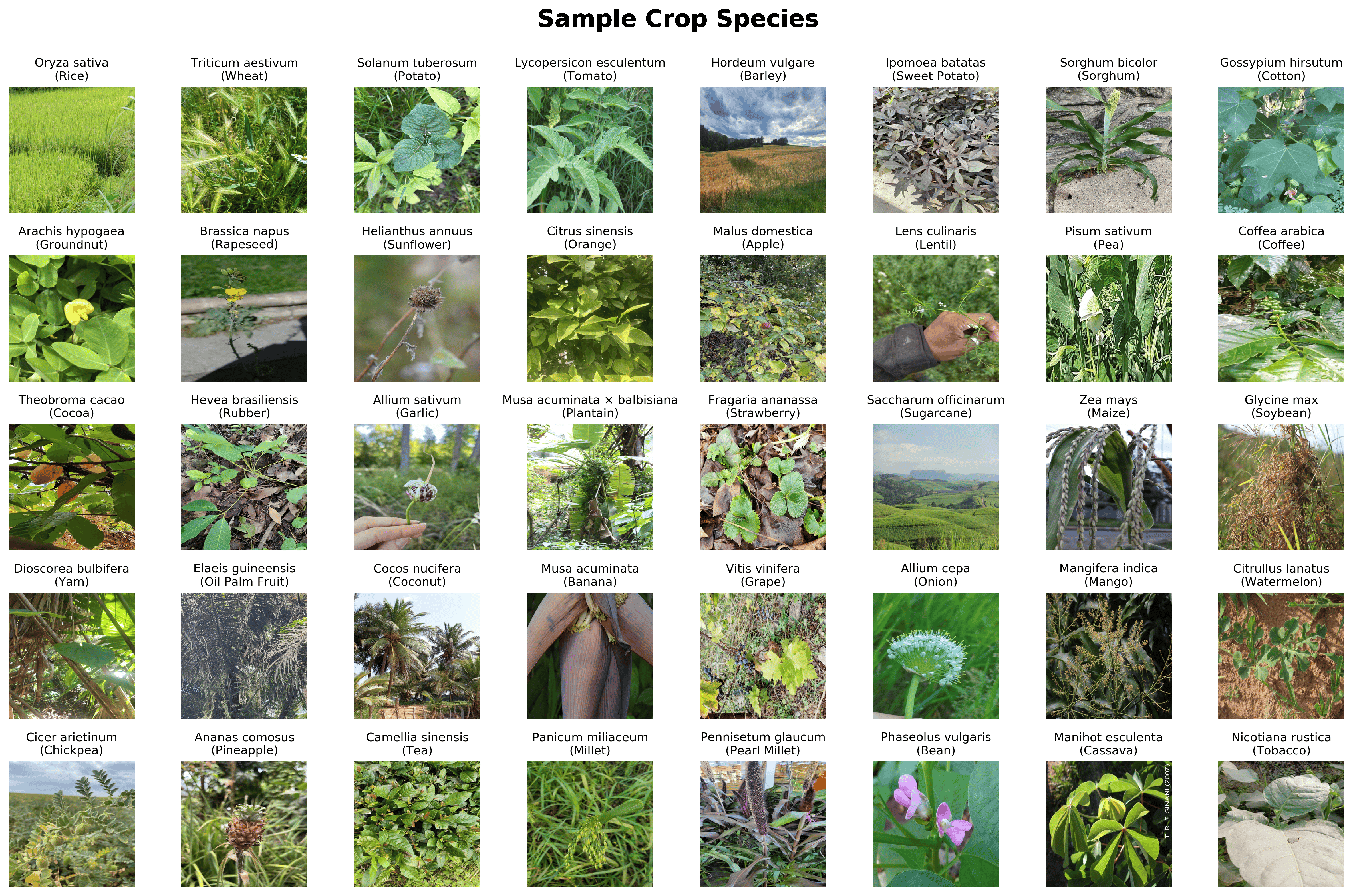}
    
    \caption{\textbf{Examples of 40 crop species from the iNatAg dataset.} This visual snapshot highlights variation in leaf shape, color, and background due to real-world, user-generated data.}
    \label{fig:crop_sample}
\end{figure*}

Fine-grained plant species classification is critical for applications in agriculture, ecology, and biodiversity monitoring. Specifically, accurate recognition supports crop health assessment, invasive species control, and precision farming workflows \cite{plantSpeciesMader, mac2019, milioto2018real, sa2018weednet}. Recent advances in deep learning and the availability of large-scale annotated datasets have enabled promising results in automated species identification \cite{mohanty2016using, ferentinos2018deep}. However, real-world field conditions introduce significant challenges: images are captured under natural lighting, with occlusions, cluttered backgrounds, and varying viewpoints \cite{ubbens2017deepplant, milioto2018real}. Moreover, many species -- particularly weeds -- exhibit especially subtle morphological differences from crops, making early-growth-stage classification especially difficult \cite{yang2020finegrained, milioto2018real}.

While deep learning models have achieved state-of-the-art results on general-purpose plant datasets, existing benchmarks often fall short in agricultural settings. PlantDoc \cite{plantdoc} focuses on plant disease identification but is limited in species diversity, containing only 2,598 images across 13 species and 27 classes, and comprising only 17 disease and 10 healthy categories. DeepWeeds \cite{olsen2019deepweeds} provides high-quality imagery of eight weed species from Australia (17,509 images), but lacks broader taxonomic coverage and does not include crop species. In order to enable broader experiments and accurate, generalized classification models, datasets of larger scales of magnitude are necessary -- particularly for generalization across classes and to distinct environments. In addition, most existing datasets support single-task classification only — typically predicting species labels without considering whether the species is agriculturally relevant, or whether it behaves as a crop or a weed \cite{Steininger2023, ilyas2025cwd30, Goncalves2023CVPR}. In practical farming scenarios, both types of predictions are necessary: one needs to know what species is present, and whether it should be preserved or removed. To address data scaling, biodiversity datasets such as iNaturalist \cite{iNat123} and PlantCLEF \cite{goeau2018plantclef} offer large-scale species coverage and rich metadata such as geolocation and taxonomy, but are not tailored for agricultural applications: they lack task-specific annotations like crop/weed labels.

To this end, AgriNet~\cite{AlSahiliAwad2022} offers a sizable collection of agricultural images (160K across 423 classes) collected from diverse geographic sources, but does not provide per-image location metadata (e.g., latitude/longitude) or multi-task labels for tasks such as crop/weed distinction. The AgML~\cite{JoshiGuevaraEarles2023} database centralizes over 50 datasets across various agricultural tasks and environments, including geographic diversity and various classification labels, but its compilation of data from distinct sources leads to the over-representation of certain species and the under-representation of others, which would lead to an unbalanced trained model. In addition, none of the data sources mentioned account for the taxonomic hierarchy present in plant species. This can lead to inherent errors in trained models, and in fact, species classification errors often occur among closely related taxa. For example, two species from the same genus may be visually similar but have vastly different implications in an agricultural field. Current evaluation metrics rarely account for taxonomic hierarchy, such as genus or family-level correctness \cite{elhamod2021hierarchy, chen2019taxonomy}.

\begin{figure*}[t] 
    \centering
    \includegraphics[width=\textwidth,height=\paperheight,keepaspectratio]{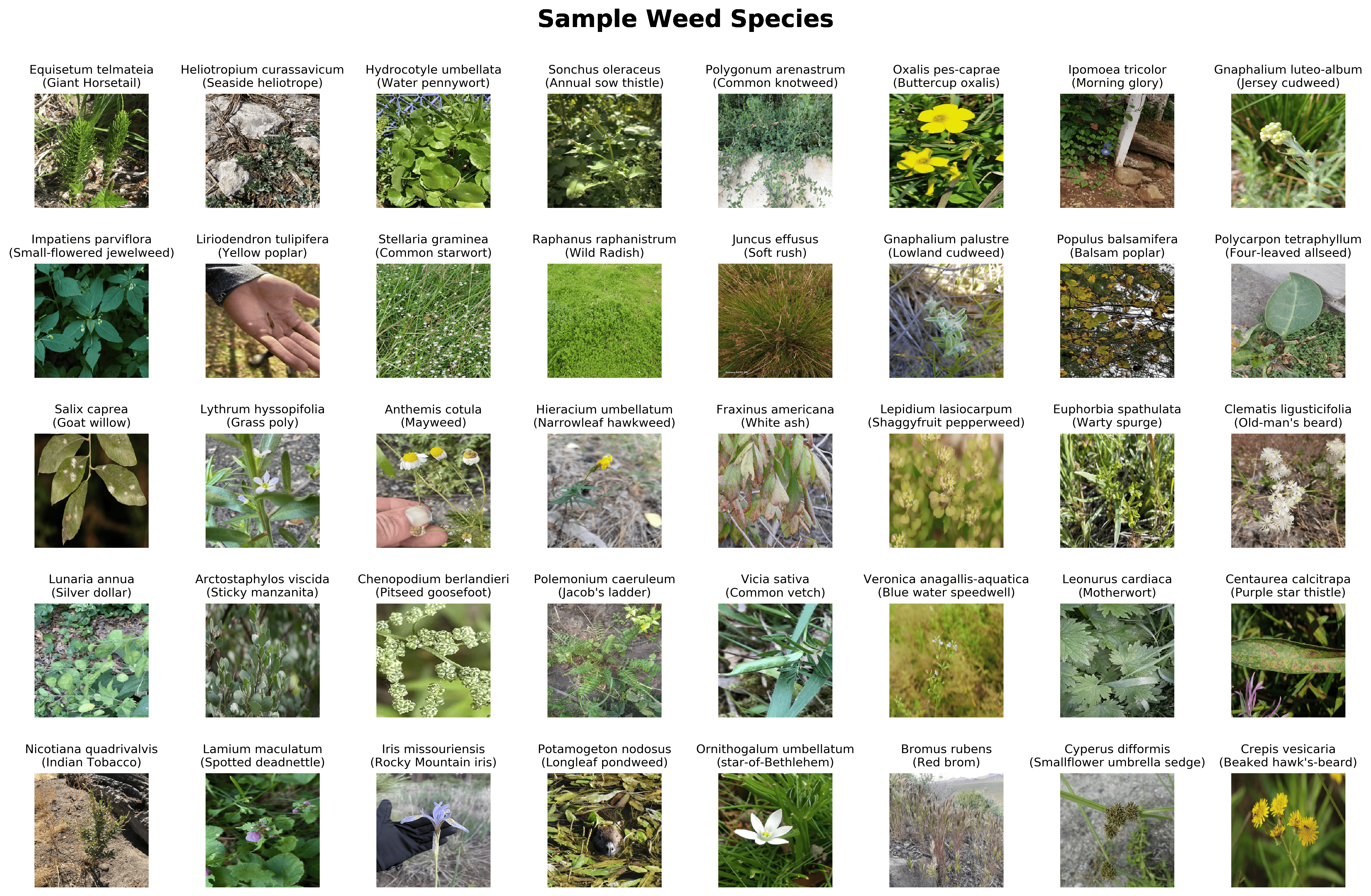}
    \caption{\textbf{Examples of 40 weed species from the iNatAg dataset.} This snapshot demonstrates the visual complexity introduced by growth stage variation, occlusion, and environmental background.}
    \label{fig:weed_sample}
\end{figure*}
\newpage

To address these gaps, we introduce iNatAg, a large-scale image dataset specifically designed for agricultural plant species classification. iNatAg is derived from publicly available data on iNaturalist \cite{inaturalist2025, inaturalistMetadata, iNat123}, but is curated through a targeted filtering process that selects only species relevant to agriculture. By cross-referencing species names with two authoritative sources -- the FAO Ecocrop database \cite{faoEcocrop} and the Weed Science Society of America (WSSA) \cite{wssa2023} composite list -- we identify and retain species known to be either crops or weeds. The resulting dataset contains 4.72 million images across 2,959 species, including 1,986 crop species and 973 weed species, covering diverse geographical regions and growing conditions. Each image in iNatAg is annotated for both species identity and crop/weed status, enabling multi-task learning. Unlike controlled lab datasets, the images in iNatAg reflect the variability of real-world, crowd-sourced data: changes in background, lighting, plant angle, and resolution. Moreover, nearly all images include latitude and longitude metadata, which we use as input features during model training. Prior work in ecological modeling has shown that incorporating spatial context can help disambiguate species with overlapping visual traits but distinct geographic distributions -- a scenario common in both invasive weed mapping and regional crop monitoring. For instance, \cite{mac2019} introduced presence-only geographical priors to improve fine-grained image classification, while \cite{cole2023sinr} proposed spatial implicit neural representations to jointly estimate species distributions at global scale.

To benchmark iNatAg, we evaluate nine configurations of Swin Transformer models \cite{liu2021swin}, across different scales (Tiny, Small, Base, Large), with and without Low-Rank Adaptation (LoRA) fine-tuning \cite{lora}, and using both image-only and image + location inputs. We assess performance across species, genus, family, and crop/weed levels using standard metrics including accuracy, precision, recall, and F1 score. We also analyze confusion matrices for commonly occurring crops and weeds, providing insight into common failure modes and the role of geolocation in improving classification outcomes.

We summarize our contributions as follows:

\begin{itemize}
    \item We introduce iNatAg, a large-scale agricultural dataset with 4.7M images spanning 2,959 crop and weed species, annotated for species and crop/weed classification.
    \item We benchmark nine Swin Transformer models across taxonomic levels and tasks, assessing the impact of LoRA fine-tuning and geospatial inputs.
    \item We find that species-level misclassifications often occur within the correct genus or family.
\end{itemize}

The dataset is publicly available through AgML, allowing researchers to easily filter, download, and experiment with iNatAg for a wide range of agricultural tasks.

\section{The iNatAg Dataset}

\begin{figure*}[t] 
    \centering
    \includegraphics[width=\textwidth,height=\paperheight,keepaspectratio]{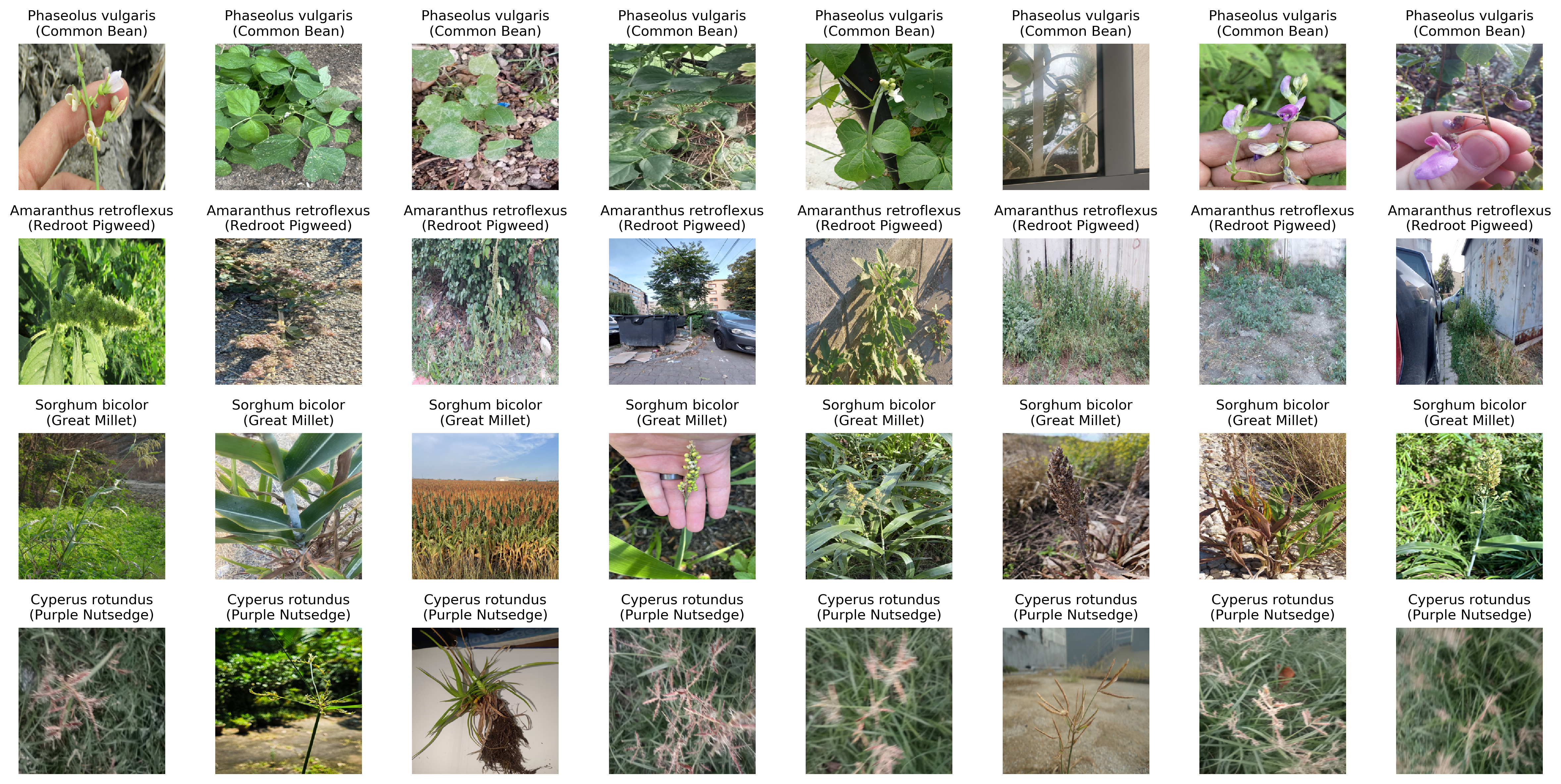}
    \caption{\textbf{Visual diversity within species and real-world crop–weed associations.} In sequential order, rows illustrate variations in Phaseolus vulgaris (crop), Amaranthus retroflexus (weed), Sorghum bicolor (crop), and Cyperus rotundus (weed).}
    \label{fig:fullpage_image0}
\end{figure*}

Studies have demonstrated the effectiveness of large agricultural datasets for training large-scale and generalized models~\cite{JoshiGuevaraEarles2023, AlSahiliAwad2022}. Building upon this work, we introduce \textbf{iNatAg}, a dataset consisting of over 4.7 million images with detailed species-level labels -- one of the largest collections of curated and standardized agricultural image data.

\subsection{Dataset Development}

The iNatAg dataset is derived from the iNaturalist dataset \cite{inaturalist2025, inaturalistMetadata}, which originally contains 269,609 plant species. To compile a comprehensive list of agricultural species, we sourced data from two key external databases. The FAO/Ecocrop database \cite{faoEcocrop} provided an extensive collection of crop species cultivated worldwide. Similarly, the Weed Science Society of America (WSSA) \cite{wssa2023} offered a well-established reference for known weed species. Using the RapidFuzz library \cite{rapidfuzz}, a high-performance string matching toolkit based on Levenshtein distance to quantify the similarity between scientific names, scientific names were mapped from iNaturalist to these external species lists. This process was essential to ensure that the dataset is well-aligned with agricultural classification tasks. Furthermore, our approach follows the broader use of fuzzy matching in taxonomic name reconciliation, such as the TaxaMatch algorithm developed for biological databases \cite{rees2014taxamatch}. Through this process, 3,529 species were identified that could be confidently categorized as either crop or weed. 

After selecting the species, we additionally curated the dataset to ensure sufficient representation of each species in terms of image availability. We extracted up to 2,500 images per species to provide a diverse yet balanced dataset. To address class imbalance and prevent over-representation of dominant species, we excluded species with fewer than 50 images and capped each species at a maximum of 2,500 images, similar to prior efforts in agricultural datasets such as DeepWeeds~\cite{olsen2019deepweeds}, where long-tailed species distributions led to performance degradation on minority classes. This step was critical to maintaining a high-quality dataset that could effectively support model training. Following this refinement, the final iNatAg dataset consists of 2,959 species, with a split of 1,986 crop species and 973 weed species. The dataset includes a total of 4,720,903 images, making it one of the most extensive datasets available for agricultural species classification.

\begin{figure*}[t] 
    \centering
    \includegraphics[width=\textwidth,height=\paperheight,keepaspectratio]{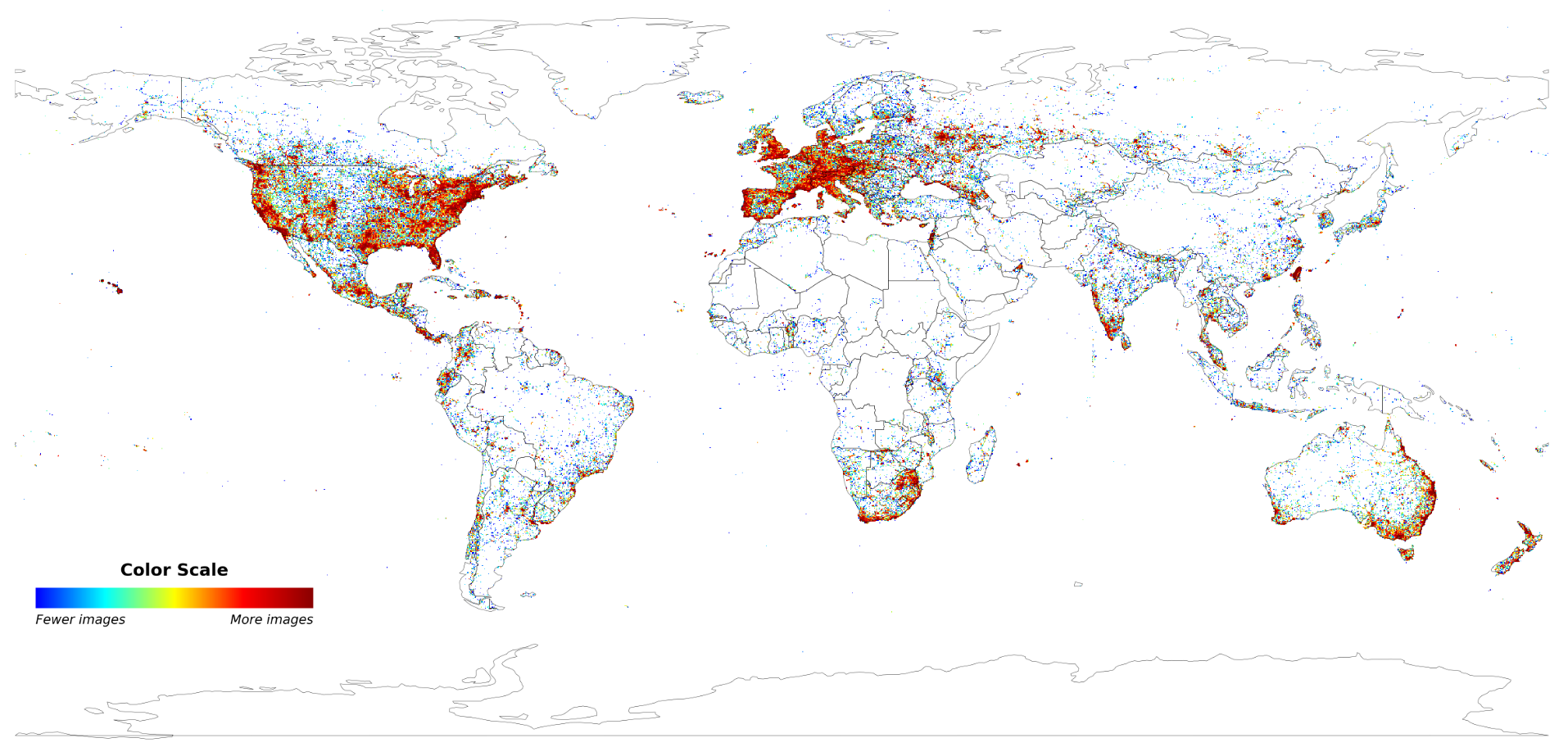}
    \caption{\textbf{Global density map of the iNatAg dataset.} Colors represent a continuous density scale, with dark red indicating more images and blue indicating fewer images. The iNatAg dataset spans multiple continents, reflecting broad ecological and geographic diversity.}
    \label{fig:fullpage_image1}
\end{figure*}

\subsection{Dataset Diversity and Expansiveness}

Figures~\ref{fig:crop_sample} and~\ref{fig:weed_sample} illustrate the wide range of species present in the iNatAg dataset, offering a visual snapshot of their diversity. Differences in leaf shape, color, and background context reflect the variation introduced by real-world conditions and user-generated data. This variety plays an important role in training models that can adapt to visual noise and field-level variability, as demonstrated in prior datasets such as PlantDoc \cite{plantdoc} and DeepWeeds \cite{olsen2019deepweeds}. Figure~\ref{fig:fullpage_image0} highlights the diversity of images within individual species in the iNatAg dataset. Each row represents a single species, and the layout also reflects real-world crop–weed associations. For example, Phaseolus vulgaris (Common Bean), a crop often coexists with Amaranthus retroflexus (Redroot Pigweed), a weed known to reduce legume yields through direct competition \cite{mirshekari2010redroot}. Similarly, Sorghum bicolor (Great Millet), a widely cultivated cereal crop, is frequently affected by Cyperus rotundus (Purple Nutsedge), a major weed that interferes with crop growth and nutrient availability \cite{hussain2021sorghum}.



Figure~\ref{fig:fullpage_image1} shows the global density map of the iNatAg dataset, generated using latitude and longitude information available for each image. The colors represent the relative density of images — dark red shows areas with more images compared to others, while blue shows areas with few or no images. The dataset covers a wide geographic range, with high-density contributions from all of North America, Europe, and Australia, and additional representation from South America, Africa, and Asia. This broad distribution brings in a variety of environmental conditions, farming practices, and species appearances, helping the model generalize better and reducing bias from any single region \cite{plantSpeciesMader}. This allows iNatAg to serve as a foundational dataset for agricultural models in any region of the world and significantly boosts its value as a dataset for improving model generalizability across environments.

\subsection{Accessing iNatAg via AgML}


To promote reproducibility and broad accessibility, iNatAg is released as part of the AgML dataset collection, with support for filtering by species, genus, or family and direct data loading through a streamlined API. The following code allows you to list available data within the iNatAg dataset:

\begin{lstlisting}[language=Python, basicstyle=\ttfamily\small]
# List iNatAg species
print(
    agml.data.public_data_sources(
        parent_dataset='iNatAg'
    )
)
\end{lstlisting}

In line with our vision of providing data across the entire taxonomic hierarchy, we also provide the ability to load data specifically according to family or species, or even common name:

\begin{lstlisting}[language=Python, basicstyle=\ttfamily\small]

# Load a collection by family names
loader = agml.data.AgMLDataLoader.from_parent(
    "iNatAg", 
    filters={"family_name": ["...", "..."]}
)

# Load by common names
loader = agml.data.AgMLDataLoader.from_parent(
    "iNatAg", 
    filters={"common_name": "..."}
)
\end{lstlisting}

In addition, we develop a subset of the iNatAg dataset for smaller-scale applications, which we call \texttt{iNatAg-mini}. It contains 560,844 images, created by sampling up to 200 images per species from the full iNatAg dataset. It is also available through the AgML with identical filtering functionality, using \texttt{iNatAg-mini} instead of \texttt{iNatAg}. These datasets can then be used in standard agricultural machine learning workflows, enabling the development of extensive applications using this data.



\section{Experiments}\label{sec:methods}

We conduct a set of experiments which are enabled by the scale of the iNatAg dataset in order to demonstrate its potential for agricultural tasks. Utilizing the species-level and crop/weed classification labels, we train Swin Transformer models to (1) identify specific plant species, (2) identify broader plant genus and families, and at the highest level, (3) distinguish crops and weeds. We also build upon our base models -- utilizing techniques such as LoRA fine-tuning and incorporation of geospatial data -- for the goal of providing insight into methods for improving agricultural classification performance.

\subsection{Model Architecture \& Training Strategy}



We used the Swin Transformer~\cite{liu2021swin}, a state-of-the-art model for multi-task learning, to perform multi-class species and genius classification as well as binary crop/weed classification. We explored the effectiveness of four model sizes: Swin Large, Base, Small, and Tiny. Models were trained using either image-only inputs or a combination of image and normalized geolocation features (latitude and longitude). Furthermore, we also validated the effectiveness of LoRA fine-tuning. In total, nine distinct training configurations were evaluated as shown in Table~\ref{tab:training_configs}.

\paragraph{LoRA Fine-Tuning.} 

We explored the use of LoRA fine-tuning \cite{lora} as an approach to improve model performance as part of our transfer learning procedure. LoRA has demonstrated potential in agricultural machine learning tasks, such as wheat field segmentation, where it enabled out-of-distribution generalization while fine-tuning only 0.7\% of model parameters \cite{zahweh2024wheat}. In our case, LoRA helped adapt large Swin Transformer models for species and crop/weed classification on iNatAg, while keeping memory usage and training overhead manageable. LoRA fine-tuning was applied to the query, key, and value layers with rank 32, alpha 16, and dropout 0.1. Residual adaptation was also enabled, allowing LoRA to integrate with the base architecture more effectively. This setup allowed us to experiment with larger model variants without exceeding typical GPU limits, enabling efficient multi-task fine-tuning across the full dataset.

\paragraph{Geospatial Features.}

Geographic coordinates were also included as auxiliary inputs to the model, allowing it to learn region-specific patterns and improve predictions for species that may look similar but occur in different locations. This is especially important in our setting, where many crops are grown in climate-specific regions and certain weeds have localized distributions. Prior work has shown that incorporating spatial context enhances accuracy in both ecological classification tasks and remote sensing applications \cite{chu, fauvel}, and recent efforts in biodiversity monitoring and geospatial AI have emphasized the role of location metadata in improving model performance~\cite{sun}. We normalize latitude and longitude coordinates, and pass them through a fully connected layer to produce a 32-dimensional embedding. This embedding is concatenated with the image features extracted by the transformer backbone. The combined representation is fed into two classification heads: one for species and one for crop/weed classification.

\subsection{Data Preprocessing \& Augmentation}

The final dataset used for training consists of 2,059 species, with a breakdown of 1,234 crop species and 825 weed species, totaling 2,059,000 images. To maintain balance and control dataset size, a maximum of 1,000 images per species was included in the dataset before splitting. The dataset was split into training, validation, and test sets as follows: 75\% (1,544,250 images) for training, 15\% (308,850 images) for validation, and 10\% (205,900 images) for testing. Each image is labeled for species classification and crop/weed classification, enabling a multi-task learning approach~\cite{ruder2017overviewmultitasklearningdeep}. We also apply various data preprocessing techniques~\cite{shorten2019survey, krizhevsky2012imagenet}, including augmentations, in order to improve the generalizability of our models. We resized our images to $384 \times 384$ for the Swin Large and Base models, and $224\times 224$ for the Swin Small and Tiny models~\cite{liu2021swin}. In addition, the following data augmentations were applied to improve data diversity in the \textit{described} way: 

\vspace{1mm}
\begin{enumerate}
    \item Random horizontal flipping \textit{(to enhance robustness)}  
    \item Random rotations \textit{(to introduce viewpoint variations)}  
    \item Color-based jittering for brightness, contrast, and saturation \textit{(to simulate variations in lighting and image capture methods)}
    \item Normalization using the ImageNet standard \textit{(in line with standard transfer learning practices~\cite{he2016deep})}
    Normalization using ImageNet mean  and standard deviation , following standard transfer learning practices  
    
\end{enumerate}
\vspace{1mm}


We also normalized latitude and longitude values, replacing any missing or invalid coordinates with an empty value. We mapped descriptive species labels to integer indices for the multi-class species classification task, and created binary weed (0) and crop (1) labels for the binary crop/weed classification task.


\subsection{Training Objective}


We trained our models to minimize the following objective:

\begin{equation}
\mathcal{L} = 0.8 \mathcal{L}_{\text{species}} + 0.2\mathcal{L}_{\text{crop+weed}}
\end{equation}

where $\mathcal{L}_{\text{species}}$ represents the multi-class loss on the species classification task, and $\mathcal{L}_{\text{crop+weed}}$ represents the binary loss on the crop/weed classification task. We use cross-entropy loss for both task losses, and $\mathcal{L}_{\text{species}}$ is weighted higher than $\mathcal{L}_{\text{crop+weed}}$  due to its larger and more complex label space. In our experiments, this weighting combination produces the best results.


\begin{table*}[h]
    \centering
    \caption{Performance of Swin Transformer variants with and without LoRA on species, genus, family, and crop/weed classification}
    \label{tab:training_configs}
    \resizebox{\textwidth}{!}{
    \begin{tabular}{llcccccccc  cc}
        \toprule
        \textbf{Model} & \textbf{Inputs}  & \multicolumn{4}{c}{\textbf{Species (\%)}} & \textbf{Genus (\%)} & \textbf{Family (\%)} & \multicolumn{4}{c}{\textbf{Crop / Weed (\%)}}\\
        \cmidrule(lr){3-6} \cmidrule(lr){7-8} \cmidrule(lr){9-12}
         &  & Accuracy \(\uparrow\) & Precision \(\uparrow\) & Recall \(\uparrow\) & F1 Score \(\uparrow\) & Accuracy \(\uparrow\) & Accuracy \(\uparrow\) & Accuracy \(\uparrow\) & Precision \(\uparrow\) & Recall \(\uparrow\) & F1 Score \(\uparrow\)\\
        \midrule
        Swin Large\textsuperscript{LoRA} & \texttt{IMAGE, GEO}  & \textbf{79.40} & \textbf{80.53} & \textbf{80.32} & \textbf{79.99} & \textbf{89.83} & \textbf{94.07} & 91.04 & 92.35 & 93.61 & 92.98\\
        Swin Large\textsuperscript{LoRA} & \texttt{IMAGE}   & 77.50 & 78.81 & 78.39 & 78.01 & 88.47 & 93.26 & 89.33 & 91.34 & 91.64 & 91.49\\
        Swin Large & \texttt{IMAGE, GEO} & 77.43 & 79.06 & 78.23 & 77.67 & 88.31 & 93.03 & 92.26 & \textbf{93.66} & 93.87 & 93.77\\
        Swin Base\textsuperscript{LoRA} & \texttt{IMAGE, GEO} & 77.11 & 78.45 & 78.09 & 77.74 & 87.73 & 92.59 & 87.66 & 90.26 & 89.85 & 90.06 \\
        Swin Base  & \texttt{IMAGE, GEO} & 77.29 & 78.76 & 78.10 & 77.64 & 87.88 & 92.65 & \textbf{92.38} & 92.92 & \textbf{95.06} & \textbf{93.97} \\
        Swin Small\textsuperscript{LoRA} & \texttt{IMAGE, GEO}   & 64.02 & 66.90 & 66.20 & 65.75 & 75.93 & 83.22 & 79.85 & 83.90 & 83.50 & 83.70\\
        Swin Small & \texttt{IMAGE, GEO}  & 69.38 & 72.01 & 71.01 & 70.52 & 81.15 & 87.52 & 88.41 & 91.45 & 90.06 & 90.75\\
        Swin Tiny\textsuperscript{LoRA} & \texttt{IMAGE, GEO} & 60.36 & 63.35 & 62.73 & 62.29 & 72.19 & 80.14 & 76.47 & 79.77 & 82.26 & 81.00 \\
        Swin Tiny & \texttt{IMAGE, GEO}  & 70.02 & 72.10 & 71.29 & 70.90 & 81.26 & 87.52 & 87.82 & 89.08 & 91.57 & 90.31\\
        \bottomrule
    \end{tabular}
    }
    
    \vspace{0.5ex}
    \footnotesize{\textsuperscript{LoRA} indicates models fine-tuned using LoRA. \texttt{IMAGE} denotes image inputs, and \texttt{GEO} indicates geospatial data (longitude/latitude).}
    \vspace{0.5ex}
    \hrule

\end{table*}

\subsection{Implementation Details}

All models were trained using the PyTorch~\cite{pytorch} library with multi-GPU support enabled via DataParallel. The AdamW optimizer \cite{weightdecay} was used with an initial learning rate of 1.5e-4 and a weight decay of 1e-3. We selected a batch size of 96. Training was conducted for up to 15 epochs, with early stopping applied if validation loss did not improve for 4 consecutive epochs. 





\subsection{Model Evaluation \& Inference}

All trained models were evaluated on a test set consisting of 205,900 images. Evaluation included both fine-grained species classification and binary crop/weed classification, with metrics such as accuracy, precision, recall, and F1 score. To assess taxonomic generalization, genus and family-level accuracy were computed by mapping species predictions to their respective higher-order groups using a predefined lookup table. Confusion matrices were also generated for the ten most common crop and weed species to further analyze misclassification patterns.

\section{Results and Discussion}

We now demonstrate the capabilities of our described models on the iNatAg dataset, referencing existing baselines where possible and demonstrating the effect of the techniques described in Section~\ref{sec:methods}. Table~\ref{tab:training_configs} provides a comparative evaluation of our various model architectures -- with and without LoRA fine-tuning -- and across species, genus and family level, and crop and weed classification. We first discuss the impact of design features including model size, fine-tuning techniques, and the inclusion of geospatial metadata (\cref{ssec:result_design}), followed by a more detailed analysis of performance specifically with relevance to the three agricultural tasks we test (\cref{ssec:result_levels}). Finally, we highlight high-level observations and discuss the role of iNatAg for future agricultural deep learning models (\cref{ssec:result_discussion}).

\subsection{Design Analysis}\label{ssec:result_design}

Following our experiments, we evaluate some of our model design choices to assess their impact on performance. 

\paragraph{Model Architecture.}

We observe that in general, larger Swin Transformer architectures perform better across all tasks. In fact, our Swin Large variants outperform their corresponding Swin Small variants by over 10\% on the species classification tasks. Larger models incorporate not only additional architectural blocks -- enabling them to learn more feature patterns in images -- but also operate on larger images. Specifically, Swin Large uses $384\times384$ images, while Swin Small and Tiny use $224\times224$ images: an over 70\% increase in image size for the former. Agricultural imagery is full of high-detail features, and distinguishing between two highly similar species requires a higher degree of detail than the broader task of simply classifying crops and weeds. This analysis is confirmed by analyzing the delta in performance between Large and Small/Tiny models on each of the three tasks: while the gap in performance is relatively low for the crop/weed classification task (only $\sim$3\%), it increases for the family classification task to roughly 8\% and as high as 15\% on the species classification task.


\vspace{-2mm}

\paragraph{LoRA Fine-Tuning.}

LoRA fine-tuning significantly improves performance for large models but is less effective -- or even detrimental -- for smaller ones. The Swin Large model with LoRA achieves the highest species accuracy at 79.40\%, outperforming the same model without LoRA at 77.43\%. In contrast, Swin Small with LoRA achieves only 64.02\%, compared to 69.38\% without LoRA. Swin Tiny shows a similar drop as it achieves 60.36\% with LoRA. 
These results are even more dramatic in the crop/weed classification model, where there is a nearly 10\% drop in accuracy as a result of LoRA fine-tuning in Swin Small and Tiny. These trends suggest that LoRA is most effective when applied to high-capacity models that can leverage the additional adaptation layers, but may hurt generalization in smaller, resource-constrained architectures.

\begin{figure}[t]
    \centering
    \subcaptionbox{\textbf{Confusion Matrix for 10 Most Common Crop Species}. The model performs well overall, with some misclassifications likely explained by morphological overlap or similarity in growth stage.\label{fig:conf_matrix1}}{%
        \includegraphics[width=\columnwidth]{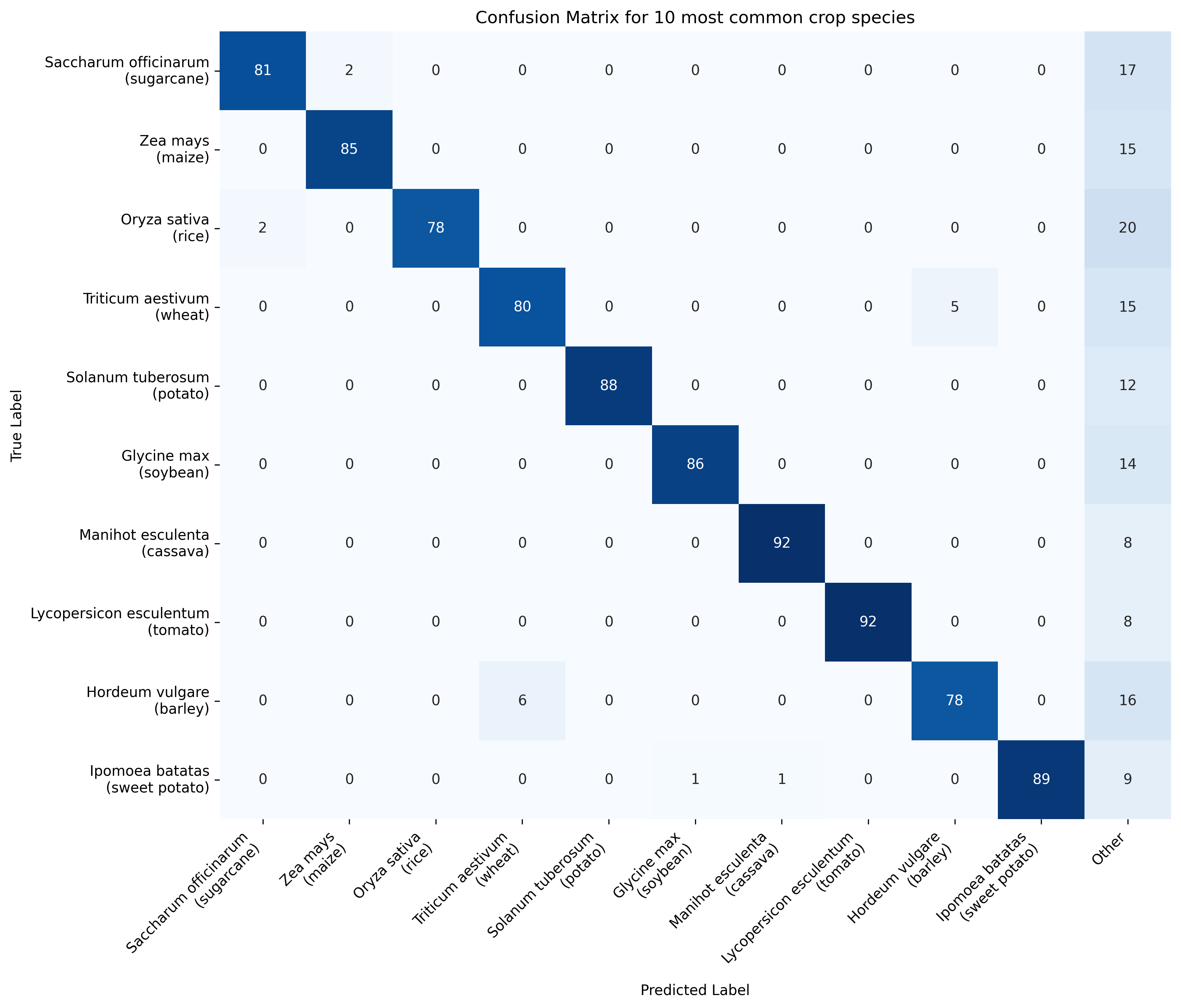}%
    }
    
    \vspace{2em} 
    
    \subcaptionbox{\textbf{Confusion Matrix for 10 Most Common Weed Species}. The model performs well overall, with the exhibited misclassifications being likely explained by a lack of distinctive visual traits and higher intra-class variability across samples.\label{fig:conf_matrix2}}{%
        \includegraphics[width=\columnwidth]{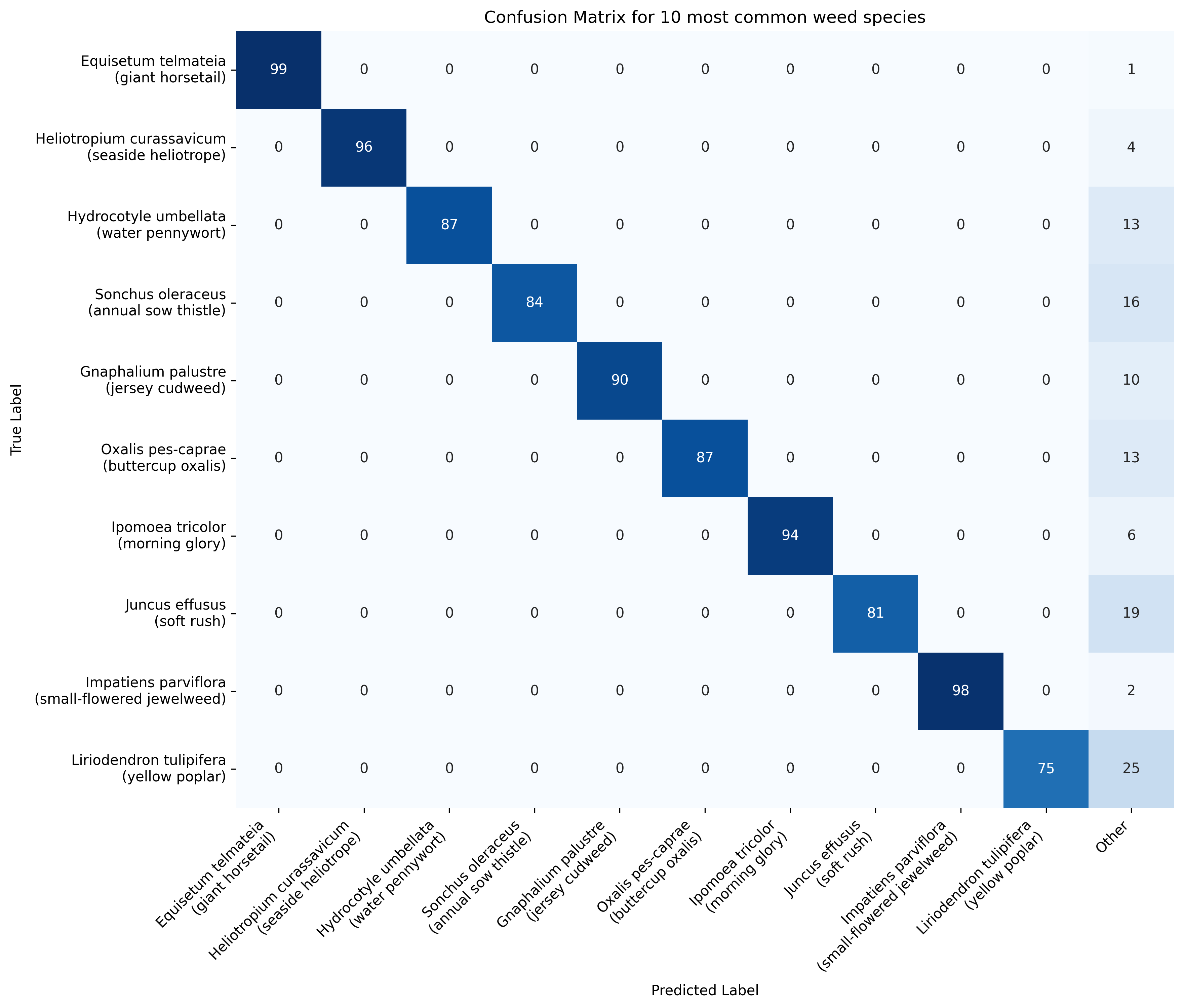}%
    }
    \caption{Comparison of confusion matrices for crops and weeds.}
\end{figure}

\paragraph{Incorporation of Geospatial Information.}

Incorporating geospatial latitude and longitude data improves classification accuracy. For example, in our best model, species accuracy increases from 77.50\% with images alone to 79.40\% with geospatial data. This improvement suggests that geolocation helps classify species that may look similar but occur in different regions. Geospatial inputs are especially useful in agricultural and biodiversity settings, where species are often tied to specific geographic zones \cite{iNat123}.

\subsection{Performance Across Classification Levels}\label{ssec:result_levels}

\paragraph{Performance Across Taxonomic Levels.}

Our results clearly indicate differences in performance relative to the taxonomic classification which models are trained on. Specifically, classification accuracy broadly increases at higher taxonomic levels. Models trained on the crop/weed task, the highest level with simple binary labels, exhibit a maximum classification accuracy of 92.38\%, while our best performing model on the species classification task, the most fine-grained task with over 2,000 labels,  reaches 79.40\%. This is complemented by the observation that our best model at the genus level achieves 89.83\% classification accuracy, while at the family level it achieves 94.07\% accuracy. This same pattern of relative performance emerges across all model sizes and architectural variations, which provides a basis for our determination that many species-level misclassifications remain within the same broader taxonomoic group.

Interestingly, we observe that architectural changes are not necessarily consistent across tasks in their performance benefits. Furthermore, as discussed in our model architecture analysis in~\cref{ssec:result_design}, the performance gap between larger and smaller models decreases at higher taxonomic levels. As a result, differences in performance between models which have less size difference, such as Swin Large and Swin Base or Swin Small or Swin Tiny, are barely 1\% for the crop/weed classification task. Factors such as random variation during training can thus serve as an explanation for variations in performance such as the Swin Base model outperforming Swin Large on the crop/weed classification task. In general, our best-performing models also have balanced precision and recall (alongside a similar F1 score), suggesting that they minimize both false negatives and false positives and provide reliable predictions. We continue this assessment now by analyzing our model's correct and mispredictions.

\paragraph{Performance on Common Crop and Weed Species.}

Our model demonstrates strong species-level performance across the 10 most common crop classes. Species like Manihot esculenta (Cassava), Lycopersicon esculentum (Tomato), etc. are classified with over 88\% accuracy, as shown in Figure~\ref{fig:conf_matrix1}. While certain species such as Rice (Oryza sativa) and Barley (Hordeum vulgare) exhibit occasional confusion, this can likely be explained by morphological overlap and growth-stage similarity. Our model also performs well on the 10 most common weed species, achieving over 90\% classification accuracy for species such as Equisetum telmateia (Giant Horsetail), Impatiens parviflora (Small-Flowered Jewelweed), etc as shown in Figure~\ref{fig:conf_matrix2}. Certain misclassifications are once again observed -- namely, Liriodendron tulipifera (Yellow Poplar) shows the most notable confusion, with 25\% of its samples classified as "Other" instead of its correct category. This can likely be explained by a lack of distinctive visual traits and higher intra-class variability across samples.


\subsection{Broader Observations and Discussion}\label{ssec:result_discussion}

We observe consistent trends across tasks and taxonomic levels: classification performance improves at broader levels, and many species-level errors remain within the correct genus or family. This aligns with prior work advocating for taxonomy-aware evaluation, including \cite{elhamod2021hierarchy, chen2019taxonomy}. While iNatAg is derived from iNaturalist data, our curated subset focuses exclusively on agriculturally relevant species, with added crop/weed annotations and multi-task supervision. This makes it distinct from biodiversity-oriented benchmarks like \cite{goeau2018plantclef, iNat123}, where models are typically trained only for species recognition. Despite these added constraints, our models achieve competitive species-level accuracy while also supporting more practical downstream tasks. We also find that including simple geolocation features -- latitude and longitude -- consistently improves performance, reinforcing insights from prior work on spatial priors and species distribution modeling \cite{mac2019, cole2023sinr}. Together, these results position iNatAg as a robust, scalable benchmark for fine-grained agricultural classification grounded in real-world variability.

\section{Conclusion}

We introduce iNatAg, a large-scale benchmark dataset for fine-grained classification of crop and weed species. iNatAg contains over 4.7 million images across 2,959 species, alongside detailed labels through the taxonomic hierarchy from species to crop and weed labels, as well as location metadata -- making it one of the largest single collections of agricultural image data to-date. We produce benchmarks using Swin Transformer-based methods alongside architectural adaptations such as LoRA fine-tuning and incorporation of geospatial metadata. Our benchmark models achieve 79.4\% accuracy at the species level, 89.83\% at the genus level, 94.07\% at the family level, and 92.38\% for crop/weed classification -- reinforce the value of spatial context and multi-task supervision in real-world agricultural settings, and demonstrating the importance of a dataset like iNatAg.
Enabled by the scale of iNatAg, we also conduct a deeper analysis into the successes and failures of these existing state-of-the-art models, making observations such as misclassifications occurring specifically within the same genus or family. Our results show the promise of iNatAg to support future research in fine-grained classification, taxonomy-aware modeling, and practical applications of agricultural AI such as precision agriculture and sustainable farming.



{
    \small
    \bibliographystyle{ieeenat_fullname}
    \bibliography{main}
}


\end{document}